\documentclass{article}
\usepackage{arxiv}
\usepackage[utf8]{inputenc} 
\usepackage[T1]{fontenc}    
\usepackage{hyperref}       
\usepackage{url}            
\usepackage{booktabs}       
\usepackage{amsfonts}       
\usepackage{nicefrac}       
\usepackage{microtype}      
\usepackage{lipsum}		
\usepackage{graphicx}
\usepackage{doi}

\usepackage{xcolor}
\usepackage{subcaption}
\usepackage{mathtools}
\usepackage{algorithm}

\usepackage{tabularx,ragged2e}
\usepackage[noend]{algpseudocode}
\usepackage{array}
\newcolumntype{P}[1]{>{\centering\arraybackslash}p{#1}}

\usepackage{verbatim}
\usepackage{diagbox}

\newtheorem{theorem}{Theorem}[section]

\newtheorem{definition}{Definition}[theorem]

\title{Sig-Wasserstein GANs for Time Series Generation}

\author{
    Hao Ni\\
    University College London\\
    London, UK\\
    \texttt{h.ni@ucl.ac.uk}\\
	\And
    Lukasz Szpruch\\
    University of Edinburgh\\
    Edinburgh, UK\\
    \texttt{l.szpruch@ed.ac.uk}\\
    \And
    Marc Sabate-Vidales\\
    University of Edinburgh\\
    Edinburgh, UK\\
    \texttt{m.sabate-vidales@sms.ed.ac.uk}\\
    \And
    Baoren Xiao\\
    University College London\\
    London, UK\\
    \texttt{baoren.xiao.18@ucl.ac.uk}\\
    \And
    Magnus Wiese\\
    University of Kaiserslautern\\
    Kaiserslautern, Germany\\
    \texttt{wiese@rhrk.uni-kl.de}\\
    \And 
    Shujian Liao\\
    University College London\\
    London, UK\\
    \texttt{shujian.liao.18@ucl.ac.uk}
}

\renewcommand{\shorttitle}{Sig-Wasserstein GANs for Time Series Generation}

\begin{document}
\maketitle

\begin{abstract}
Synthetic data is an emerging technology that can significantly accelerate the development and deployment of AI  machine learning pipelines. In this work, we develop high-fidelity time-series generators, the SigWGAN, by combining continuous-time stochastic models with the newly proposed signature $W_1$ metric. The former are the Logsig-RNN models based on the stochastic differential equations, whereas the latter originates from the universal and principled mathematical features to characterize the measure induced by time series.  SigWGAN allows turning computationally challenging GAN min-max problem into supervised learning while generating high fidelity samples.  We validate the proposed model on both synthetic data generated by popular quantitative risk models and empirical financial data. Codes are available at \href{GitHub}{https://github.com/SigCGANs/Sig-Wasserstein-GANs.git}
\end{abstract}

\keywords{generative modelling, neural networks, expected signature, log-signature, rough path theory, Wasserstein generative adversarial networks}
\section{Introduction}

The ability to model time-series data accurately is critical for numerous applications in the finance industry.  In particular, synthetically generated time-series datasets can facilitate training and validation of data-driven risk models and enable data sharing by respecting the demand for privacy constraints. We refer the reader to  \cite{assefagenerating,bellovin2019privacy} for the overview of the applications and challenges for synthetic data generation and to  \cite{cuchiero2020generative, Buehler2020data, Koshiyama, gierjatowicz2020robust, Buehler2019, QuantGAN} for generative modelling perspective of some of the classical problems in quantitative risk management. While generative modelling has been highly successful in generating samples from seemingly high dimensional probability measures, off-the-shelf techniques, such as generative adversarial network’s (GAN) \cite{Goodfellow}, struggle to capture the temporal dependence of joint probability distributions induced by time-series data. Furthermore, the min-max objective function of classical GANs make them notoriously difficult to tune. In this paper, we use mathematically principled feature extraction machinery that emerged from the theory of rough paths to reduce the min-max formulation of GANs to an optimization problem. The proposed method, called Sig-Wasserstein GAN, can handle irregularly spaced data streams of variable lengths and is poetically efficient for data sampled at high frequency.

\paragraph{Related work}
It is by now well documented that popular machine learning frameworks enhanced with path signatures achieve the state of the art performance across many time series learning tasks. For example, the combination of rough path theory and variational autoencoder showed the strength in simulating financial time series in small data environment \cite{Buehler2020data}.  In \cite{arribas2020sig} authors showed that by combining signatures with classical quantitative finance models provide novel perspective on neural SDEs and lead to efficient calibration procedure. Related results have been obtained in \cite{kidger2020neural}. The most related to this work is \cite{ni2020conditional}. There authors developed the conditional Sig-Wasserstein GAN to simulate time series that mimics the conditional law of the future time series given the past. This work focuses on the unconditional case instead. Our Logsig-RNN generator extends the work of \cite{ liao2021logsig},  as we do not require the equal time dimension of input data and output data. 

The article is structured as follows. In Section 2 we overview the key elements of rough path theory. In Section 3 we present classical GAN framework which we then extend to Sig-Wasserstein GAN in Section 4. Section 5 contains numerical examples. 



\section{Rough path theory}
When working with time-series data, especially sampled at high frequency and/or with irregular time stamps, it is useful to take a continuous-time perspective and view data as unknown continuous-time dynamics samples.  In particular, rough path theory offers a mathematically principled and universal way of describing the continuous-time data trajectories (paths), which in turn allows designing computationally efficient algorithms. We briefly summarise these ideas below.    
\subsection{Signature of a path}
Let $E:=\mathbb{R}^d$ and $J$ be a compact time interval. Let $X: J \to E$ denote a continuous path endowed with a norm denoted by $|\cdot|$. We first introduce the \emph{$p$-variation} as a measure of the roughness of a path. The larger $p$-variation is, the rougher a path is.

\begin{definition}[$p$-variation] Let $p \geq 1$ be a real number. Let $X: J \to E$ be a continuous path. The $p$-variation of $X$ on the interval $J$ is defined by
$
\|X\|_{p,J}=\sup_{\mathcal{D}\in J}\left[\sum_{j=0}^{r-1}|X_{t_{j+1}}-X_{t_j}|^p\right]^{\frac{1}{p}}
$, where the supremum is taken over any finite time partition of $J$, i.e. $\mathcal{D}=(t_1,t_2,\cdots,t_r).$ 
\end{definition}

\begin{definition}[The signature of a path]
Let $X : J \rightarrow E$ be a continuous path of ﬁnite $p$-variation such that the following integration makes sense. The signature of $X$ denoted by $S(X)$ is deﬁned as an infinite series of $X^{k}_{J}$, i.e. $ S(X)_J=(1,X^1_J,\cdots,X^k_J,\cdots) ,$
where $$X^k_J= \mathop{\int\cdots\int}_{u_1<\cdots <u_k;u_1,\cdots,u_k \in J} dX_{u_1} \otimes \cdots \otimes dX_{u_k} , \forall k\geq1.$$
Let $S_k(X)_J$ denote the truncated signature of $X$ of degree $k$, i.e. $S_k(X)_J=(1,X^1_J,...,X^k_J).$
\end{definition}

\paragraph{Path augumentations} There are a few commonly used path augumentations methods to accompany with the signature feature to retain good properties of the signature.  In our work, we use three augumentation methods (a) Time augumentation, (b) Visiability transformation, and (c) Lead-lag transformation, which add the extra feature dimension to encode the information on the time stamps, the starting point of the path and the lagged process respectively. We refer \cite{morrill2020generalised} for the precise definition of the above path augmentations. For ease of notation, we denote the space of the time-augmented and visibility transformed paths of finite $p$-variation by $\Omega^{p}_{0}(J, E)$. Such augumented path has the path dimension $2d+1$ where $d$ is the dimension of the original path. 

Intuitively the signature of a path plays a role of a non-commutative polynomial on the path space.
With appropriate path augmentations, the signature of a path has the \emph{uniqueness} and \emph{universality}, which make the signature a useful candidate for the feature set of a path: \\
\textbf{Uniqueness}: The signature of a path determines the path up to tree-like equivalence \cite{hambly2010uniqueness,boedihardjo2015uniqueness}.
More specifically, when restricted the path space to $\Omega^1_0(J, \mathbb{R}^d)$, the signature map is bijective.
In other words, the signature of an augumented path by time augmentation and visibility transformation determines the path completely.\\
\textbf{Universality}:  Any continuous functional on the paths in $\Omega^1_0(J, \mathbb{R}^d)$ can be arbitrarily well approximated by a linear functional of truncated signatures when the degree of the signature is high enough. To state it more precisely we have:
\begin{theorem}\label{th sig}
Consider a compact set $K \subset \Omega^{1}_{0}(J, \mathbb{R}^{d})$. Let $f: K \to \mathbb{R}$ be any continuous function. Then, for any $\epsilon>0$, there exists an integer $M>0$, and a linear functional $l \in T^{(M)}(\mathbb{R}^{2d+1})^*$ acting on the truncated signature of degree $M$ such that
\begin{equation}
    \sup_{X \in K}|f(X) - \langle l, S_{M}(X)\rangle| < \epsilon
\end{equation}
\end{theorem}

\subsection{Log-signature of a path}
The log-signature is a parsimonious representation for the signature feature. To define the log-signature, we introduce the logarithm of an element in the tensor algebra space, i.e. $T((E)) = \oplus E^{\otimes n}$. Let $a=(a_0,a_1,\cdots,a_n)\in T((E))$ be such that $a_0=1$ and $t=a-1.$ Then the logarithm map is defined as follows:
\begin{equation}
    \log(a)=\log(1+t)=\sum_{n=1}^{\infty}\frac{(-1)^{n-1}}{n}t^{\otimes n}, \forall a\in T((E)).
\end{equation}
\begin{definition}
The log signature of path $X$ is the logarithm of the signature of the path $X$, denoted by $LogSig(X)$. Let $LogSig_k(X)$ denote the truncated log signature of a path $X$ of degree $k$.
\end{definition}

\textbf{Uniqueness:} Like the signature, the log-signature has the uniqueness as there is one-to-one correspondence between the signature and the log-signature.\\
\textbf{Dimension reduction}
    When truncated by the same degree, the log-signature is of lower dimension compared with the signature feature in general. It be used for significant dimension reduction when the path dimension $d$ and the truncated degree of the (log)-signature $k$ is large. See the comparison figure of the dimension of the signature and log-signature.
{
\begin{table}[H]
\centering
\resizebox{0.8\textwidth}{!}{
\begin{tabular}{|c|rl|rl|rl|rl|rl|rl|}
\hline
\backslashbox{$k$~}{$d$}  & \multicolumn{2}{c|}{2} & \multicolumn{2}{c|}{3} & \multicolumn{2}{c|}{4} & \multicolumn{2}{c|}{5} & \multicolumn{2}{c|}{6} & \multicolumn{2}{c|}{7} \\ \hline
1 & 3          & \textbf{2}         & 4          & \textbf{3}         & 5          & \textbf{4}         & 6          & \textbf{5}         & 7         & \textbf{6}         &     8       &  \textbf{7}         \\
2 & 7          & \textbf{3}        & 13         & \textbf{6}         & 21         & \textbf{10}        & 31         &\textbf{15}        & 43        & \textbf{21}        &        57    &      \textbf{28}     \\
3 & 15         & \textbf{5}         & 40         & \textbf{14}        & 85         & \textbf{30}        & 156        & \textbf{55}        & 259       & \textbf{91}        &     400       &      \textbf{140}    \\
4 & 31         & \textbf{8}         & 121        & \textbf{32}        & 341        & \textbf{90}        & 781        & \textbf{205}       & 1555      & \textbf{406}      &        2801    &    \textbf{728}   \\
5 & 63         & \textbf{14}        & 364        &\textbf{80}        & 1365       & \textbf{294}       & 3906       & \textbf{829}       & 9331      & \textbf{1960}      &        19608    &  \textbf{4088}         \\ \hline
\end{tabular}}
\caption{The left integer is the signature dimension whereas the right integer (in bold) is the log-signature dimension. }
\end{table}
}

\subsection{Expected signature of a stochastic process}
Let us consider a $E$-valued stochastic process $X$ under the probability space. Assume that the signature of $X$ is well defined almost surely, and $S(X)$ has finite expectation. We call $\mathbb{E}[S(X)]$ the expected signature of $X$. Intuitively the expected signatures serves the moment generating function, which can characterize the law induced by a stochastic process under some regularity condition. More concretely, an immediate consequence of Proposition 6.1 in \cite{chevyrev2016characteristic} on the uniqueness of the expected signature is summarized in the below theorem:

\begin{theorem}
Let $X$ and $Y$ be two $\Omega_{0}^{1}(J, E)$-valued random variables. If $\mathbb{E}[S(X)] = \mathbb{E}[S(Y)]$ and $\mathbb{E}[S(X)]$ has the infinite radius of convergence, then $X = Y$ in the distribution sense.\end{theorem}

\section{Wassersetein Generative Adversarial Network (WGAN)}
Let $(\Omega,\mathcal{F}, \mathbf{P})$ be a probability space, under which $\mu$ and $\nu$ are two distributions induced by $\mathcal{X}$-valued stochastic process. The Kantorovich-Rubinstein dual representation of Wassersetein-1 ($W_{1}$) metric defines a distance between two measures $\mu$ and $\nu$, denoted by $ W_{1}(\mu, \nu)$ as follows:  

\begin{equation}\label{eq:W1_dual}
     W_{1}(\mu, \nu)= \sup_{\left\Vert f\right\Vert_{\text{Lip}}\leq 1 }  \mathbb{E}_{X\sim \mu}[f(X)]-\mathbb{E}_{X\sim \nu}[f(X)],
\end{equation} 
where the supremum is over all the 1-Lipschitz functions $f : \mathcal{X} \rightarrow \mathbb{R}$ with its Lipschitz norm smaller than 1, i.e. for all $x_1, x_2 \in \mathcal{X}$, $|f(x_{1})-f(x_{2})|\leq |x_{1}-x_{2}|$.

In the context of generative modelling, let $\mathcal{Z}$ denote a latent space and $Z \in \mathcal Z$ denote a variable with the known distribution $\mu_{Z} \in \mathcal{P}(\mathcal{Z})$. Let $\mu \in \mathcal{P}(\mathcal{X})$ denote a target distribution of observed data. The aim of Wassersetein Generative Adversarial Network (WGAN) is to train a model that induces distribution $\nu$, so that $W_{1}(\mu,\nu)$ is small. The WGAN is composed of the generator $G$ and the discriminator $D$. 
 The generator $G_{\theta}: \mathcal{Z} \rightarrow \mathcal{X}$ is a parameterized map transporting the latent distribution $\mu_{Z}$ to the model distribution $\nu$, i.e. the distribution induced by $ G_{\theta}(Z)$, where $\theta \in \Theta^{g}$ is a parameter set. To discriminate between real and synthetic samples, one parameterises the test function $f$ in definition of $W_1$ metric (Eqn. \eqref{eq:W1_dual}) by a network $f_{\eta}$ with the parameter set $\eta \in \Theta^{d}$. Training the generator entails solving min-max problem. Indeed,  to find optimal $(\theta^{\star}, \eta^{\star})$, one needs to solve
\begin{eqnarray*}
\min_{\theta} \max_{\Vert f_\eta\Vert_{\text{Lip}} \leq 1} \mathbb{E}[f_{\eta}(X)] - \mathbb{E}[f_{\eta}(G_{\theta}(Z))].
\end{eqnarray*}
In practice, the min-max problem is solved by iterating gradient descent-ascent algorithms and it convergence can be studied using tools from game theory \cite{mazumdar2019finding,lin2020gradient}.  It is well known that first order methods that are typically used in practice to solve the min-max problem might not converge, even if the  convex-concave case, \cite{daskalakis2018limit,daskalakis2017training,mertikopoulos2018cycles}. Consequently the adversarial training is notoriously difficult to tune, \cite{farnia2020gans,mazumdar2019finding}, and generalisation error is very sensitive to the choice of discriminator and hype-parameters as it was demonstrated in large scale study in \cite{lucic2017gans}.

\section{Sig-Wassersetein Generative Adversarial Network (Sig-WGAN) }

In this work we design algorithms for generating time-series data. Let $x_{1:T} := (x_{1}, \cdots, x_{T}) \in \mathcal X \subseteq \mathbb{R}^{d \times T}$ denotes a sample, or trajectory, of a time series data and assume that $x_{1:T}$ is distributed according to some unknown target distribution $\nu \in \mathcal P(\mathcal X)$. Unlike majority of work in the literature, we don't assume that the data points are equidistant from each other.
Given $N$ independent trajectories $(x^i_{1:T})_{i=1}^N$ sampled from $\nu \in \mathcal P(\mathcal X)$ or given one long trajectory $(x_{iT,(i+1)T})_{i=0}^N$ of stationary data, the aim of generative modelling  is to learn a model capable of producing high fidelity data samples from $\nu \in \mathcal P(\mathcal X)$, without explicitly modelling the target distribution.
\subsection{Signature Wassersetein-1 (Sig-$W_{1}$) metric}

We propose a new Signature Wasserstein-1 (Sig-$W_{1}$) metric on the measures on the path space $\mathcal{X}=\Omega^1_0(J, \mathbb{R}^d)$ by combining the signature feature and the $W_1$ metric to achieve better computation efficiency. Let $\mu$ and $\nu$ be two measures on the path space $\mathcal{X}$.
When using $W_{1}$ with. discrete time series, one needs to fix the time dimension of time series a prior. Here we (linearly) interpolate the data and compute its signature, possibly using appropriate path augmentation methods without information loss. The signature, as a universal and principled feature,  encodes the temporal information regardless of the sampling frequency and variable length.  Hence one may consider using the $W_1$ metric on the signature space to define a distance between the measure induced by two measures on the path space $\mu$ and $\nu$, i.e.
\begin{eqnarray}\label{eqn_W1_sig}
W_{1}^{\text{Sig space}}(\mu, \nu)= & \underset{\left\Vert f\right\Vert_{\text{Lip}}\leq 1 }{\sup}  \mathbb{E}_{\mathbf{X}\sim \mu}[f(S(\mathbf{X)})]-\mathbb{E}_{\mathbf{X}\sim \nu}[f(S(\mathbf{X}))] \end{eqnarray}
While the use of signature features space $W_{1}^{\text{Sig space}}$ significantly reduces the time dimension of the problem, practical challenges of solving min-max problem remains. 
By working with the signature feature, one can reduce the computation of  $W_{1}^{\text{Sig space}}$ distance over the class of Lipschitz functionals to the linear functionals on the signature space thanks to the universality property of the signature. This motivates us to consider the proposed Signature Wasserstein-1 metric between $\mu$ and $\nu$ defined by 
\begin{equation}\label{sigw1}
    \text{Sig-W}_1(\mu, \nu) := \underset{ \centering \small{L\textrm{ is linear, $\Vert L\Vert \leq 1$}}}{\sup}
\mathbb{E}_{\mathbf{X}\sim\mu}[L(S(X))]- \mathbb{E}_{\mathbf{X}\sim\nu}[L(S(X))]\,.
\end{equation}
Hence, by using  $\text{Sig-W}_1$ we reduced the nonlinear optimisation task to the linear problem with constraints. 

In practice, one needs to truncate the infinite dimensional signature to a finite degree for numerical computation of $W_{1}^{\text{Sig space}}$ and Sig-$W_1(\mu, \nu)$. The factorial decay of the signature enables us to approximate the signature in Eqn. \eqref{sigw1} by its truncated signature up to degree $M$ for a sufficiently large $M$. Therefore we propose to define the truncated Sig-$W_1(\mu, \nu)$ metric up to a degree $M$ as follows:
\begin{equation}\label{sigw1:M}
    \text{Sig-W}^{(M)}_1(\mu, \nu) := \sup_{\Vert L \Vert\leq1, L\text{ is linear}}
L(\mathbb{E}_{\mathbf{X}\sim \mu}[S_M(X)]- \mathbb{E}_{\mathbf{X}\sim \nu}[S_M(X)]),
\end{equation}

When the norm of $L$ is chosen as the $l_2$ norm of the linear coefficients of $L$, this reduced optimization
problem admits the analytic solution
\begin{equation}\label{sigw1:analytic}
    \text{Sig-W}^{(M)}_1(\mu, \nu) := |\mathbb{E}_{\mu}[S_M(X)]- \mathbb{E}_{\nu}[S_M(X)]|
\end{equation}
where $|.|$ is $l_2$ norm. In \cite{chevyrev2018signature},
if one chooses the truncated signature up to degree $M$ as the feature map, then the corresponding
Maximum Mean Discrepancy (Sig-MMD) is the square of $\text{Sig-W}^{(M)}(\mu,\nu)$.

\begin{figure}[!ht]
\includegraphics[width=1.\linewidth]{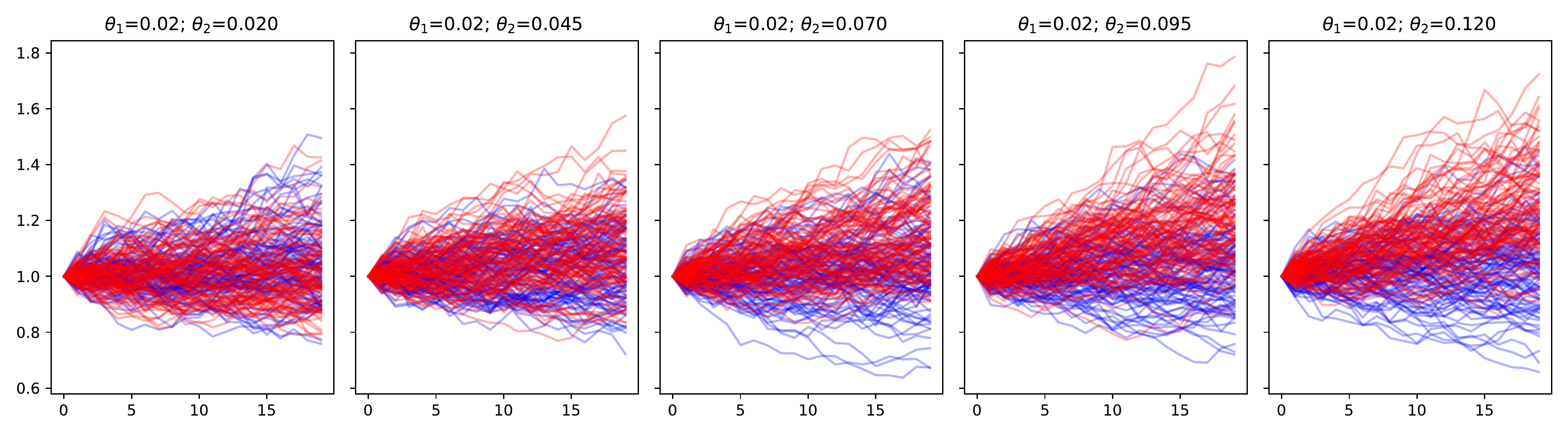}
\centering
\includegraphics[width=0.6\linewidth]{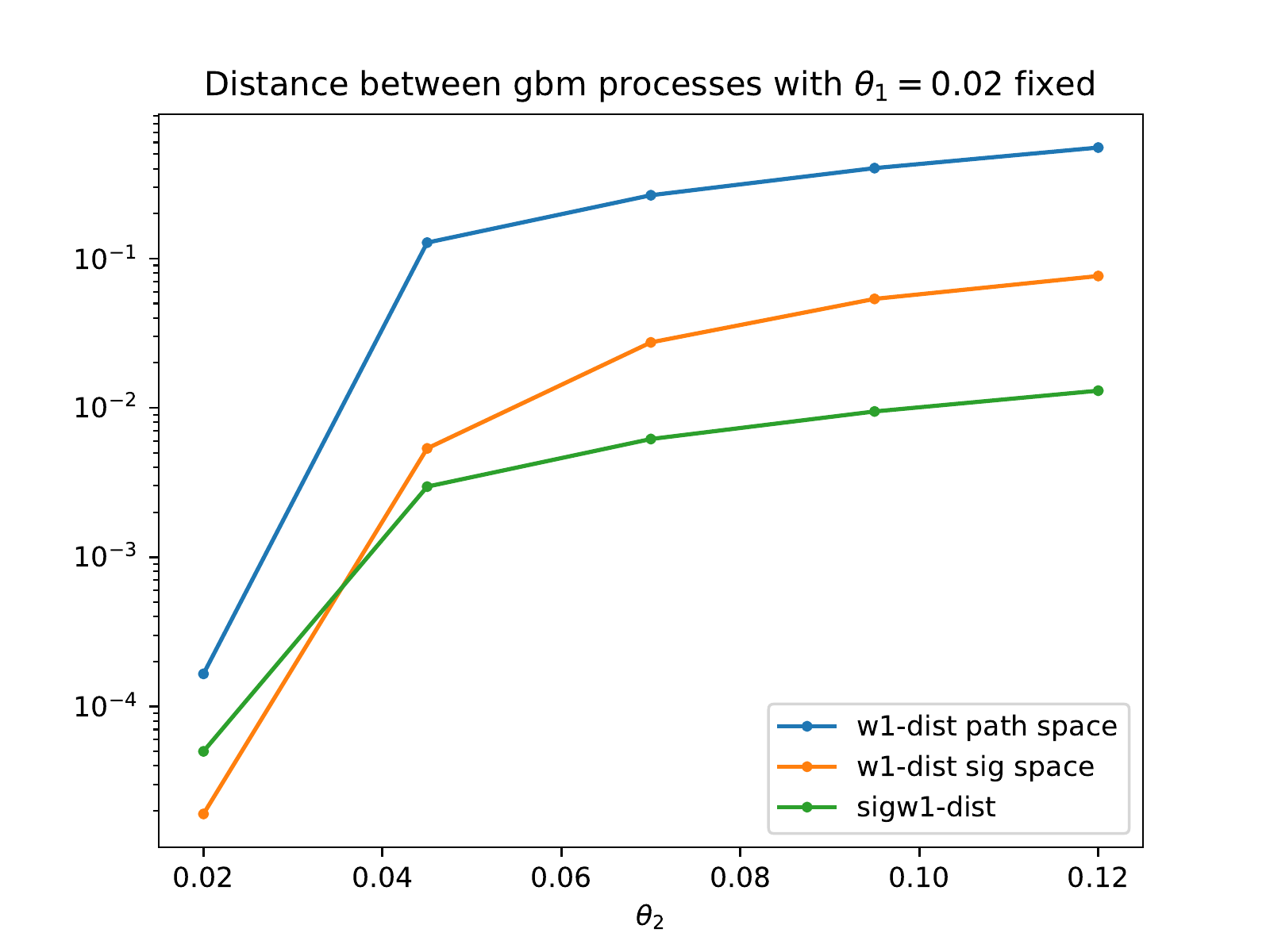}
\caption{The top row displays blue and red samples from $\mathbf X, \mathbf{\hat X}$ respectively for fixed $\theta_1$, and different values of $\theta_2$}
\label{fig:gbms}
\end{figure}

The following toy example illustrates the relationship between the Sig-$W_1$ distance and the $W1$ distance between two path distributions. Let $X=(X_t)_{t\in[0,T]}, \hat{X}=(\hat{X}_t)_{t\in[0,T]}$ two 1-dimensional GBMs given by,
\[
\begin{split}
dX_t = & \, \theta_1 X_t dt + \sigma X_t dW_t,\,\,X_0=1; \\
d\hat{X}_t = & \,  \theta_2 \hat{X}_t dt + \sigma \hat{X}_t d\hat{W}_t,\,\,\hat{X}_0=1,
\end{split}
\]
with the same volatility but with possibly different drifts $\theta_1, \theta_2$. Let $\mu,\nu$ be the laws of $X, \hat X$. We fix $\sigma=0.1, \theta_1=0.02$, and for $\theta_2 = 0.02+ j 0.025, j=0,\ldots,4$. When $\theta_2$ is increasing, the discrepancy between $X$ and $\hat X$ is increasing. We calculate three distances, i.e. $W_{1}^{\text{path space}}$, $W_{1}^{\text{Sig space}}$ and Sig-$W_1$ to quantify the distance between $X$ and $\hat{X}$ for different $\theta_2$, which all increase when enlarging $\theta_2$ as expected.
Since $\text{Sig-W}^{(M)}_1(\mu, \nu)$ admits an analytic solution, it is cheaper to calculate than $W_{1}^{\text{path space}}(\mu, \nu)$ and $W_{1}^{\text{Sig space}}(\mu, \nu)$, where one needs to parametrise $f$ by a neural network and optimise its weights. We observe in Figure~\ref{fig:gbms} how these three values increase with similar rate as $\theta_2$ increases.

\subsection{LogSig-RNN Generator}
In this section, we introduce two generators of continuous type, which both are motivated by the numerical approximation of stochastic differential equations (SDEs). The first generator is the Neural SDE, which approximates the vector field by a feed-forward neural network in the Euler scheme, while the second generator -- the Logsig-RNN resembles the high-order Taylor approximation of SDEs by combining the recurrent neural network with the log-signature. \\

Fix a filtered probability space $(\Omega, \mathcal{F}, \{\mathcal{F}_{t}\}_{t\geq 0}, \mathbf{P})$, under which $W = (W_t)_{t \in [0, T]}$ be a $d$-dimensional Brownian motion. Let $\mathbf{W} = (\mathbf{W}_{t})_{t \in [0, T]}$ denote the time-augmented Brownian motion, where $\mathbf{W}_{t} = (t, W_{t})$ for ease of the notation. Consider a $\mathbb{R}^{e}$-valued process $X = (X_{t})_{t \in [0, T]}$, which satisfies the following time-homogeneous SDEs driven by the Brownian motion $W$ with the drift term $\mu$ and the volatility term $\sigma$, i.e.
\begin{eqnarray}\label{eq:sde}
dX_{t} = \mu(X_{t}) dt + \sigma(X_{ t})dW_{t}:=f( X_{t})d\mathbf{W}_{t},
\end{eqnarray}
where the stochastic integral is taken in the stratonovich sense and the vector field $f: x \times (s, \omega) \mapsto \mu(x) s+ \sigma(x)\omega, \forall  x \in \mathbb{R}^{e}, s \in \mathbb{R} \text{ and } \omega \in \mathbb{R}^{d}.$
\footnote{The time-inhomogeneous SDE can be viewed as the projection of the solution to time-homogeneous SDE by lifting $X$ to its time-augmented process. } 


One can approximate the solution $X$ defined by \eqref{eq:sde} via the step-$M$ Taylor expansion locally; when $s$ and $t$ are close,
\begin{eqnarray}\label{sde1}
X_{t} - X_{s} \approx \sum_{m = 1}^{M} f^{\circ k}(X_s) \int_{s <s_1 <\cdots <s_m<t} d\mathbf{W}_{s_{1}} \otimes \cdots \otimes d\mathbf{W}_{s_{m}}, 
\end{eqnarray}
where $0<s<t<T$, $f^{\circ m}: \mathbb{R}^{e} \rightarrow L((\mathbb{R}^{d+1})^{\otimes m}, \mathbb{R}^{e}) $ is defined inductively by 
\begin{eqnarray*}
f^{\circ 1} = f; f^{\circ {m+1}} = D(f^{\circ m})f
\end{eqnarray*}
with $D(g)$ denoting the differential of the function $g$. This leads to the numerical approximation scheme by pasting the local step-$M$ Taylor approximation together. Indeed, fix time partitions $\Pi_{h} = (u_{j})_{j = 0}^{N_{1}}$ and $\Pi_{X} = (t_{k})_{k = 0}^{N_{2}}$ of $[0, T]$, which are time partition of Taylor expansion and the time discretization of the solution $X$ respectively. Without loss of generality, $\Pi_{h} \subset \Pi_X$. We define $\hat{X}$ evaluated at the time partition $\Pi_{X}$ inductively; let $\hat{X}_{0} = x_0$. For any $t_{k} \in \Pi_{X}$ find $j$ such that $t_{k} \in (u_{j-1}, u_{j}]$, we apply the Taylor approximation of $X_{t_{k}}$ around the reference time point $t = u_{j-1} \in \Pi_{h}$, and hence obtain
\begin{eqnarray}
\hat{X}_{t_{k}} &=& \hat{X}_{u_{j-1}} + \sum_{m = 1}^{M} f^{\circ m}(\hat{X}_{u_{j-1}}) \int_{u_{j-1} <s_1 <\cdots <s_m<t_k} d\mathbf{W}_{s_{1}} \otimes \cdots \otimes d\mathbf{W}_{s_{k}} \nonumber \\
&:=& F(\hat{X}_{u_{j-1}},S^{M}(\mathbf{W}_{u_{j-1}, t_{k}})) = \tilde{F}(\hat{X}_{u_{j-1}},\text{LogSig}^{M}(\mathbf{W}_{u_{j-1}, t_{k}})).\label{sde_logsig}
\end{eqnarray}
 
Motivated by \cite{liao2019learning}, we approximate $\tilde{F}$ in \eqref{sde_logsig} by using a recurrent neural network. This leads to  generalized Logsig-RNN model, which maps a $d$-dimensional Brownian motion $\mathbf{W}_{[0, T]}$ to $(o_{t_{k}})_{k = 1}^{N_{2}} \in \mathbb{R}^{N_2 \times e}$ as the generator of $\mathbb{R}^{e}$-valued stochastic process: $\forall j \in \{1, \cdots, N_{1}\}$ and $\forall k \in \{1, \cdots, N_2\}$, 
\begin{eqnarray} 
h_{t_{k}} &=& \sigma_1(\theta_1 h_{u_{j-1}} + \theta_2 \text{LogSig}^{M}(\mathbf{W}_{u_{j-1}, t_{k}})); \\
o_{t_k} &=& \sigma_2(\theta_{3}h_{t_k}).\nonumber
\end{eqnarray}
where $\sigma_1, \sigma_2$ are two activation functions and $\Theta = \{\theta_{1}, \theta_{2}, \theta_3\}$ is the learnable parameter set. When $\Pi_{h} = \Pi_{X}$, this is exactly the Logsig-RNN model in \cite{liao2019learning}

\begin{figure}[H]
    \centering
    \includegraphics[width = 0.9\textwidth]{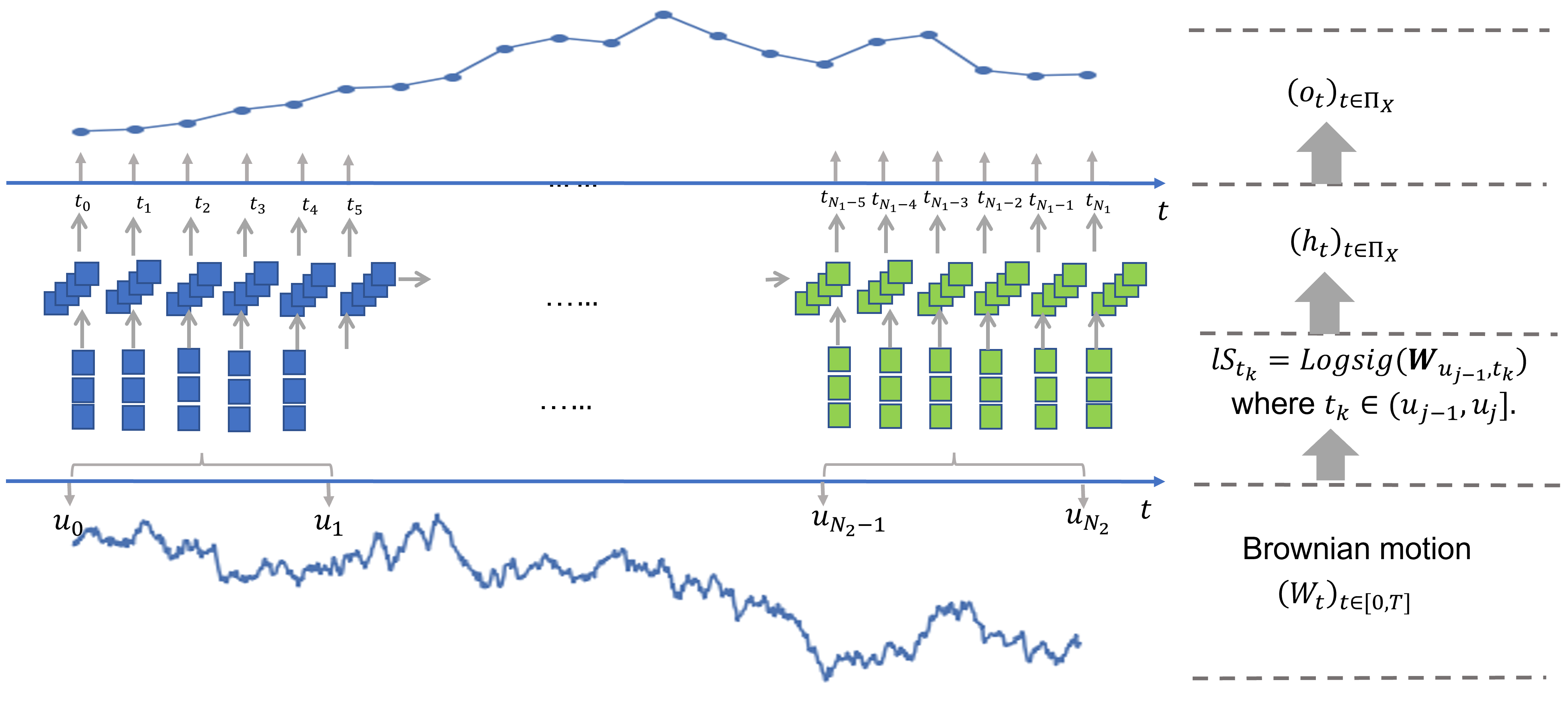}
    \caption{The pictorial illustration of the generalized Logsig-RNN model.}
    \label{fig:logsig_rnn}
\end{figure}

\section{Numerical results}
To validate the performance of the proposed Sig-Wasserstein GANs (SigWGANs), we consider three datasets, i.e. (1) synthetic data generated by multi-dimensional Geometric Brownian motion (GBM); (2) synthetic data generated by the rough volatility model; (3) stock price and realized volatility data. The former two datasets are representatives of commonly-used Markovian and non-Markovian model for the underlying price process.  For each dataset, we compare the proposed SigWGANs with WGANs to demonstrate the advantages of Sig-Wasserstein metric in terms of the accuracy and efficiency. Besides for the SigWGANs, we benchmark the proposed Logsig-RNN generators of continuous type against the Long short-term memory (LSTM) model. To assess the quality of generated data, we consider the following test metrics:
\begin{enumerate}
    \item \textbf{Sig-W$_1$ metric}.
    \item \textbf{Marginal distribution metric}. To assess the fitting of the marginal distribution, we compute the average of Wasserstein distance (also called earth mover's distance, EMD for short) between each marginal distribution $X_{t}^{(i)}$ of the real time series and fake time series over all time $t$ and feature channel $i$ as the marginal distribution metric.
    \item \textbf{Correlation metric}. To quantify the fitting of spatial and temporal dependence, we consider how close the correlation of $X_{t}^{(i)}$ and  $X_{s}^{(j)}$ for any feature coordinate $i$ and $j$ and any time $s$ and $t$. The correlation metric of $X$ and $\hat{X}$ is defined as 
   $ \text{cor}(X, \hat{X}) = \sum_{s,t = 1}^{T}\sum_{i, j = 1}^{d}  \left|\rho(X_{s}^{(i)}, X_{t}^{(j)}) - \rho(\hat{X}_{s}^{(i)}, \hat{X}_{t}^{(j)})\right|$, where $\rho(X, Y)$ denotes the correlation of two real-valued random variables $X$ and $Y$.
\end{enumerate}
The smaller test metrics indicates better performance. 

Moreover, on the synthetic dataset, we test the predictive performance of the generative model, which is trained on the dataset of one time frequency and is tested on another time frequency. The better fitting performance on the test time frequency shows the robustness and generalization of the trained generative model against variation in sampling time.  
\subsection{Multi-dimensional Geometric Brownian motion (GBM)}
As a motivating example, we consider a $d$-dimensional GBM $X=(X_{t})_{t\in [0, T]}$ satisfying the following SDE: for $i \in \{1, \cdots, d\}$,
$$\label{sde:GBM}
dX^{(i)}_{t} = \mu_{i} X_t^{(i)} dt + \sigma_{i} X_{t}^{(i)}d\tilde{W}_{t}^{(i)}$$
where $X_{t} \in \mathbb{R}^{d}$, $\mu\in \mathbb{R}^d$, $\sigma\in \mathbb{R}^d$ and $\tilde{W}$ is a $d$-dimensional correlated Brownian motions  with correlation matrix $(\rho_{ij})_{i,j\in\{1,\cdots,d\}}$. In this example, we set $\mu^i=0.1$, $\sigma^{i}=0.2$, $\forall i\in\{1,...,d\}$ and $\rho_{ij}=0.5$, for $i\neq j$. We use the equally spaced time partition of the time interval $[0,1]$ with step size $\Delta t= 10^{-3}$, and simulate $4000$ samples of the solution $X$ to the SDE (\ref{sde:GBM}) using the analytic formula: $X_{t_{n}}^{(i)}=X_{t_{n-1}}^{(i)}\exp[(\mu_i-\frac{1}{2}\sigma_{i}^2)\Delta t+\sigma_{i} (\tilde{W}_{t_{n}}^{(i)}-\tilde{W}_{t_{n-1}}^{(i)})]$ where $t_{n} = n \Delta t$ and $i \in \{1, \cdots, d\}$. Here we specify $d =3$.
\subsubsection{Performance comparison of SigWGAN v.s WGAN}
\begin{figure}[htb]
    \centering 
\begin{subfigure}{0.8\textwidth}
  \includegraphics[width=1\linewidth]{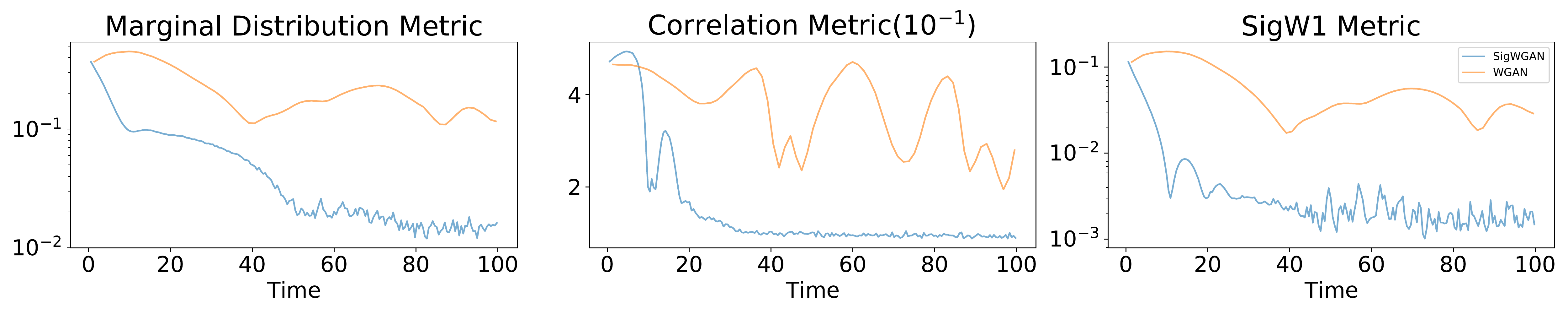}
  \caption{LogsigRNN}
  \label{fig:1}
\end{subfigure}\hfill 
\quad
\begin{subfigure}{0.8\textwidth}
  \includegraphics[width=\linewidth]{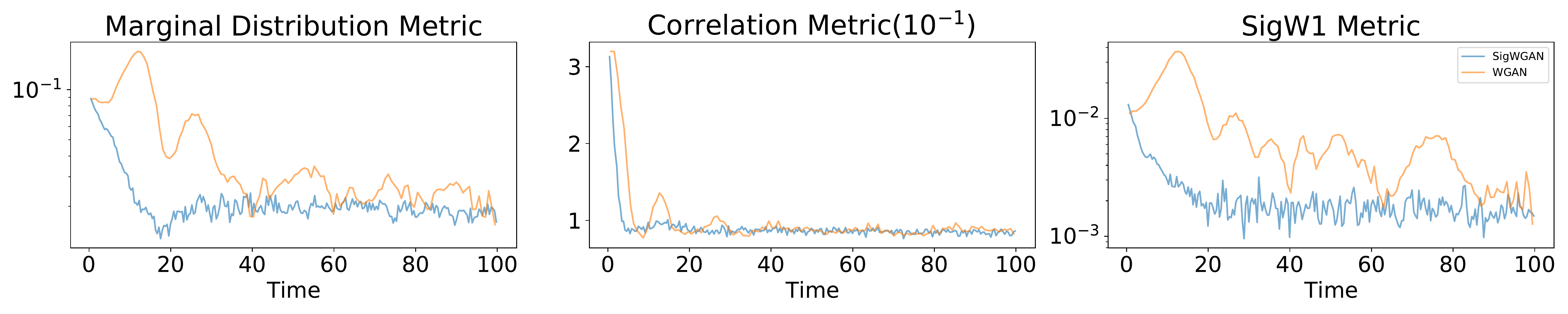}
  \caption{LSTM}
  \label{fig:2}
\end{subfigure}
\caption{The upper/lower panel are the evolution of the test metrics of the Logsig-RNN/LSTM generator during training for the first 100 seconds. From left to right, the test metrics are Marginal distribution metric (Left), Correlation metric (Middle) and SigW1 metric (Right). The results are  trained on $3$-dimensional Geometric Brownian motion. }
\label{3dimGBM_time}
\end{figure}

Figure \ref{3dimGBM_time} shows that by replacing the W1 discriminator by Sig-W1 discriminator, three test metrics drop more quickly at the beginning of training period, which indicates that the Sig-WGAN is more efficient to train. Besides, regardless of the generator, the Sig-WGAN outperforms the WGAN in terms of the three test metrics during the training. 

\subsubsection{Sensitively analysis in terms of number of time steps} Table \ref{tab:GBM_timestep}.
\begin{table}[!ht]
\centering
\resizebox{0.9 \textwidth}{!}{
\begin{tabular}{|l|c|c|c|c|c|c|c|c|}
\hline &\multicolumn{4}{c| }{W1} & \multicolumn{4}{c|}{Sig-W1 } \\ \hline
\centering{$N_T$} & $10$   & $20$  & $50$  & $100$  & $10$ &  $20$   &  $50$ &  $100$ \\ \hline
\multicolumn{9}{|c|}{SigW1 Metric(1e-1)}                     \\ \hline
LSTM    &0.140& 0.125& 0.677& 0.651 &0.096&0.064& \textbf{0.020}& 0.196\\ \hline
LogsigRNN &  0.264& 0.128& 0.162& 0.160& \textbf{0.047}& \textbf{0.040}& 0.042& \textbf{0.037}\\ \hline
\multicolumn{9}{|c|}{Correlation Metrics (1e-3)}                     \\ \hline
LSTM    & 1.057& 0.862& 5.512& 2.658&\textbf{0.478}& \textbf{0.852}& 0.9698& 3.391 \\ \hline
LogsigRNN &  2.125& 1.811& 9.277& 2.219 & 0.962& 1.032& \textbf{0.967}&\textbf{0.919}\\ \hline
\multicolumn{9}{|c|}{Marginal Distribution Metric}                            \\ \hline
LSTM    &0.052 & 0.065& 0.132& 0.082       & \textbf{0.026}&\textbf{0.043}& \textbf{0.036} &0.140\\ \hline
LogsigRNN &0.122& 0.059& 0.317& 0.067 &   0.050 & 0.047& 0.050 & \textbf{0.045}\\ \hline
\multicolumn{9}{|c|}{Training time (seconds) }                        \\ \hline
LSTM      & 376.55& 374.78& 375.44& 375.53      &    134.33 &134.66 &134.71&135.89\\ \hline
LogsigRNN &  682.21& 718.21&721.21& 781.21       & 261.23& 322.33& 371.21&472.21\\ \hline
\end{tabular}}
\caption{Evaluation for 3-dimensional GBM with various numbers of timesteps $N_T \in \{10, 20, 50, 100\}$.}\label{tab:GBM_timestep}
\end{table}
As shown in Table \ref{tab:GBM_timestep}, the Sig-W1 GAN with the LSTM generator works best when $N_{T}$ is small (e.g. $10$ or $20$). However, when increasing the number of time step $N_{T}$, it gets more challenging for the LSTM generator to learn the joint path distribution in terms of correlation metric. In contrast to the LSTM, the performance of the LogsigRNN is much robust in terms of high frequency sampling of the data. In particular, combined with Sig-W1 metric, the Logsig-RNN generator achieves the correlation metric and marginal distribution of narrow range $[0.045, 0.050]$ and $[0.919, 1.032]$ for each $N_{T} \in \{10, 20, 50, 100\}$ respectively. For $N_{T} = 100$, when using the SigWGAN, the test metrics of the LSTM generator is about four times of that of the Logsig-generator.
For a fixed generator, we observe the consistent performance improvement of using Sig-W$_{1}$ over the W$_1$ metric as the discriminator.
\subsection{Rough Volatility model}
We consider a rough stochastic volatility model for an asset price process $S_t$, which satisfies the below SDE:
\begin{equation}\label{eq:bergomi}
    \begin{split}
        dS_t = &\,\,\, \sqrt{v_t} S_t dZ_t \\
        v_t := &\,\,\, \xi(t) \exp\left(\eta W_t^H - \frac{1}{2}\eta^2 t^{2H}\right)
    \end{split}
\end{equation}
where $\xi(t)$ denotes the forward variance and $W_t^H$ denotes the fBM given by 
\[
W_t^H := \int_0^t K(t-s) dW_s, \quad K(r) := \sqrt{2H}r^{H-1/2}
\]
where $Z_t, W_t$ are (possibly correlated) Brownian motions. In our experiments, the synthetic dataset is sampled from~\eqref{eq:bergomi} with $t\in[0,1], H=0.25, \xi(t) = 0.25, \eta=0.5, \rho=0$ where $\rho dt = d\langle W, Z \rangle_t$. Each sampled path is a stream of data of 20 points sampled uniformly between $[0,1]$.  

For the training details, we train the generators to learn 
\begin{itemize}
    \item the log price distribution,
    \item the log price and the log volatility joint distributions
\end{itemize} 
using WGAN and Sig-W1 GAN. In both learning algorithms we scale the log-price paths by $2$ and the log-volatiliy by $0.5$ so that they have similar variance. In addition, in the SigWGAN we augment the paths by adding time dimension and visibility transformation before computing the expected signature up to degree $4$. 
For a fair comparison, we train the Wasserstein GAN and the Sig-W1 GAN for $2\, 500$ iterations of the generator. The Sig-W1 GAN provides an explicit form for the Sig-W1 distance between two path distributions, which implies that the overall number of gradient steps is $2\, 500$. However, training the Wasserstein GAN involves training the generator and the discriminator; in our settings, for every gradient step of the generator, we take three gradient steps on the discriminator, thus the overall number of gradient steps is $10\, 000$.

Table \ref{Table_RV_Results} shows the evaluation metrics of the trained generators.
 
\begin{table}[!ht]
\centering
\begin{tabular}{|l|c|c|c|c|}
\hline &\multicolumn{2}{c|}{W1} & \multicolumn{2}{c|}{Sig-W1 } \\ \hline
\centering{Data $(X_t)_{t}$} & $(S_t)$ ~~~~~~~~~~~  & $(S_t, v_{t})$   & $(S_t)$   & $(S_t, v_{t})$  \\ \hline
\multicolumn{5}{|c|}{SigW1 Metric}                     \\ \hline
LSTM& 0.29 & 0.42 & \textbf{0.07} & \textbf{0.17}\\
\hline
LogsigRNN& 0.20 & 1.43 & 0.11 & 0.32\\
\hline
\multicolumn{5}{|c|}{Correlation Metric(1e-3)}                     \\ \hline
LSTM& 4.76 & 5.21 & 1.58 & 8.23\\
\hline
LogsigRNN& 2.19 & 8.19 & \textbf{1.09} & \textbf{4.82} \\
\hline
\multicolumn{5}{|c|}{Marginal Distribution Metric(1e-2)}                     \\ \hline
LSTM& 2.37 & 1.95 & 1.62 & 4.83\\
\hline
LogsigRNN& 1.17 & 9.25 & \textbf{1.38} & \textbf{4.03}\\
\hline
\multicolumn{5}{|c|}{Training time(seconds)}                 \\ \hline
LSTM& \textbf{158} & \textbf{205} & 164 & 256\\
\hline
LogsigRNN& 814 & 845 & 352 & 452\\
\hline
\end{tabular}
\caption{The test metrics of the trained models on a one dimensional price data $(S_{t})_{t \in [0, T]}$ and two dimensional price and volatility data $(S_{t}, v_{t})_{t \in [0, T]}$ respectively.}\label{Table_RV_Results}
\end{table}
Sig-W1 GANs outperform GANs in all evaluation metrics, except for the training time in the case of the LSTM generator. The LSTM and the LogSigRNN generators have similar performances in terms of correlation metric, SigW1 metric and Marginal Distribution metric. 

Regarding the training time, the LSTM generator is marginally cheaper to train with the W1-GAN algorithm, even if the training of the W1-GAN involves 4x more gradient steps than the Sig-W1 GAN. This is due to the fact that calculating the signature of the augmented path in the Sig-W1 GAN algorithm accounts for the additional iterations taken to train the W1-GAN. However, for a more complex generator as is the case of the LogSigRNN, the Sig-W1 GAN is 2x cheaper than the W1-GAN in terms of training time.  

\subsubsection{Correlation metric comparison}
\begin{figure}[htb]
    \centering 
\begin{subfigure}{0.8\textwidth}
  \includegraphics[width=1\linewidth]{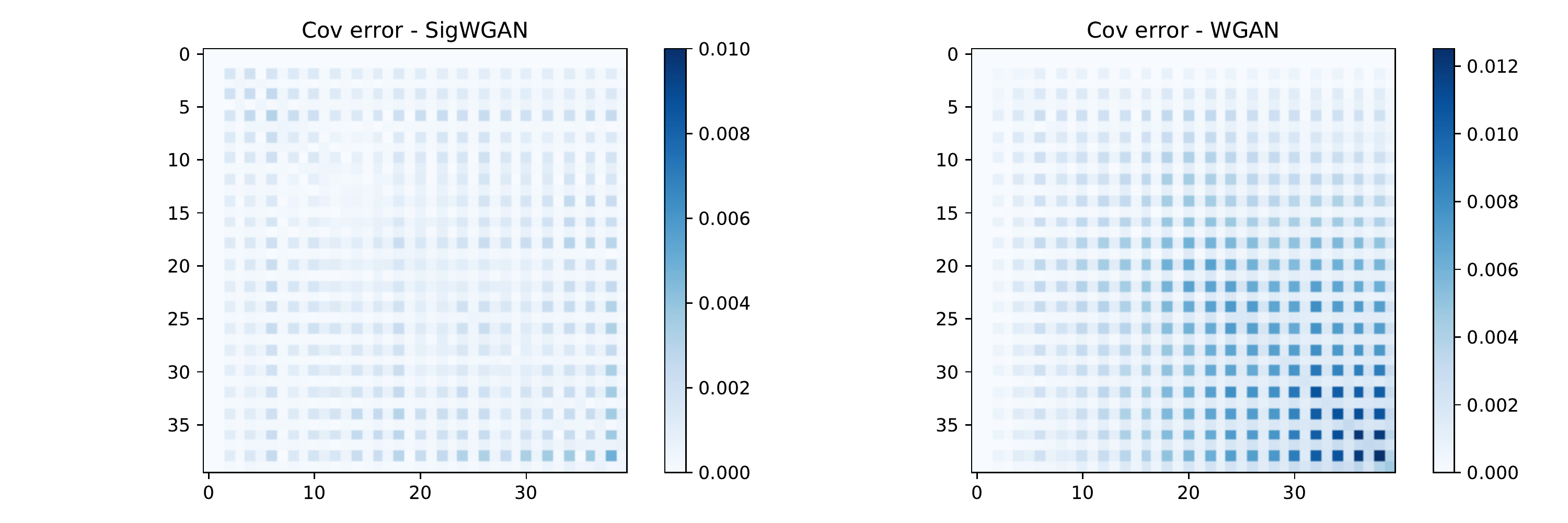}
  \caption{LogsigRNN}
  \label{fig:1}
\end{subfigure}\hfill 
\quad
\begin{subfigure}{0.8\textwidth}
  \includegraphics[width=\linewidth]{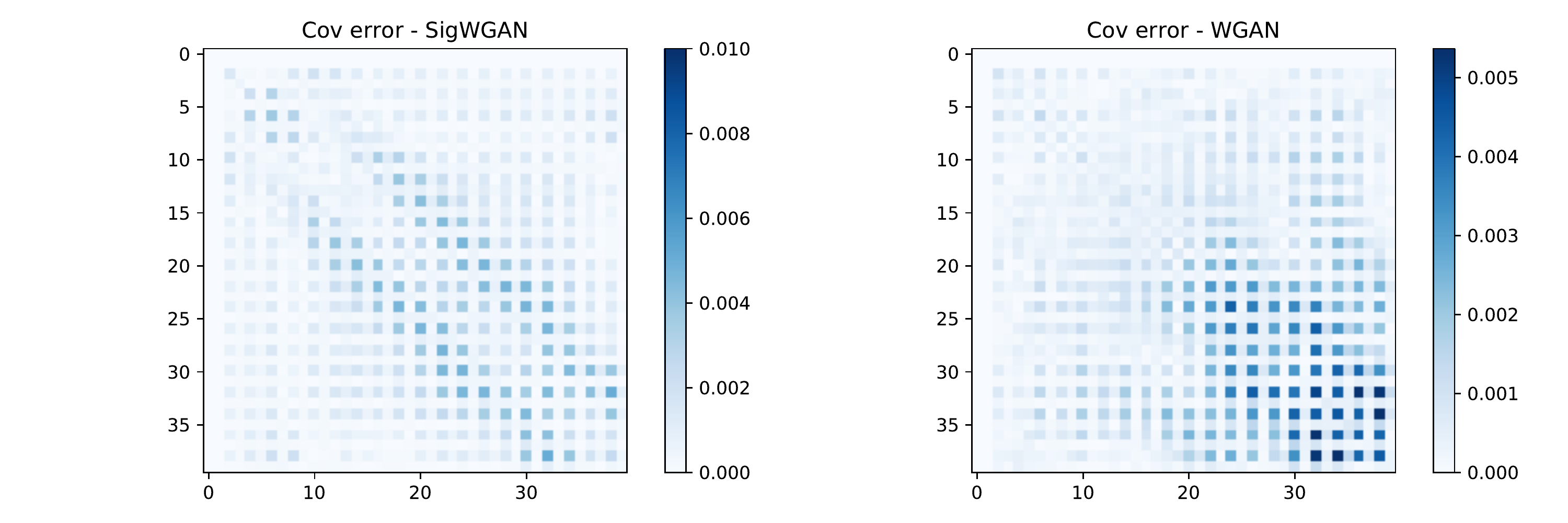}
  \caption{LSTM}
  \label{fig:2}
\end{subfigure}
\caption{The upper/lower figures are the covariance error plots of the Logsig-RNN/LSTM generators for each dimension $(S_t, v_t)$ and each timestep. From Left to Right, each heatmap displays the covariance error of the Sig-WGAN and the WGAN respectively.}
\label{fig:cov}
\end{figure}

The error plot of the covariance matrix $\left(\text{cov}(X_{s}^{(i)}, X_{t}^{(j)})\right)_{i, j \in \{1, 2\}, s, t \in \{1, \cdots, N_{T}\}}$ in Figure \ref{fig:cov} clearly shows that the combination of the Logsig-RNN and SigW1 metric are able to capture the temporal and spatial dependency of the rough volatility data best. The WGAN struggles to the temporal covariance of the volatility data. 

\subsubsection{Robustness to the variable sampling frequency}
\begin{figure}[htb]
    \centering 
\begin{subfigure}{0.3\textwidth}
  \includegraphics[width=1\linewidth]{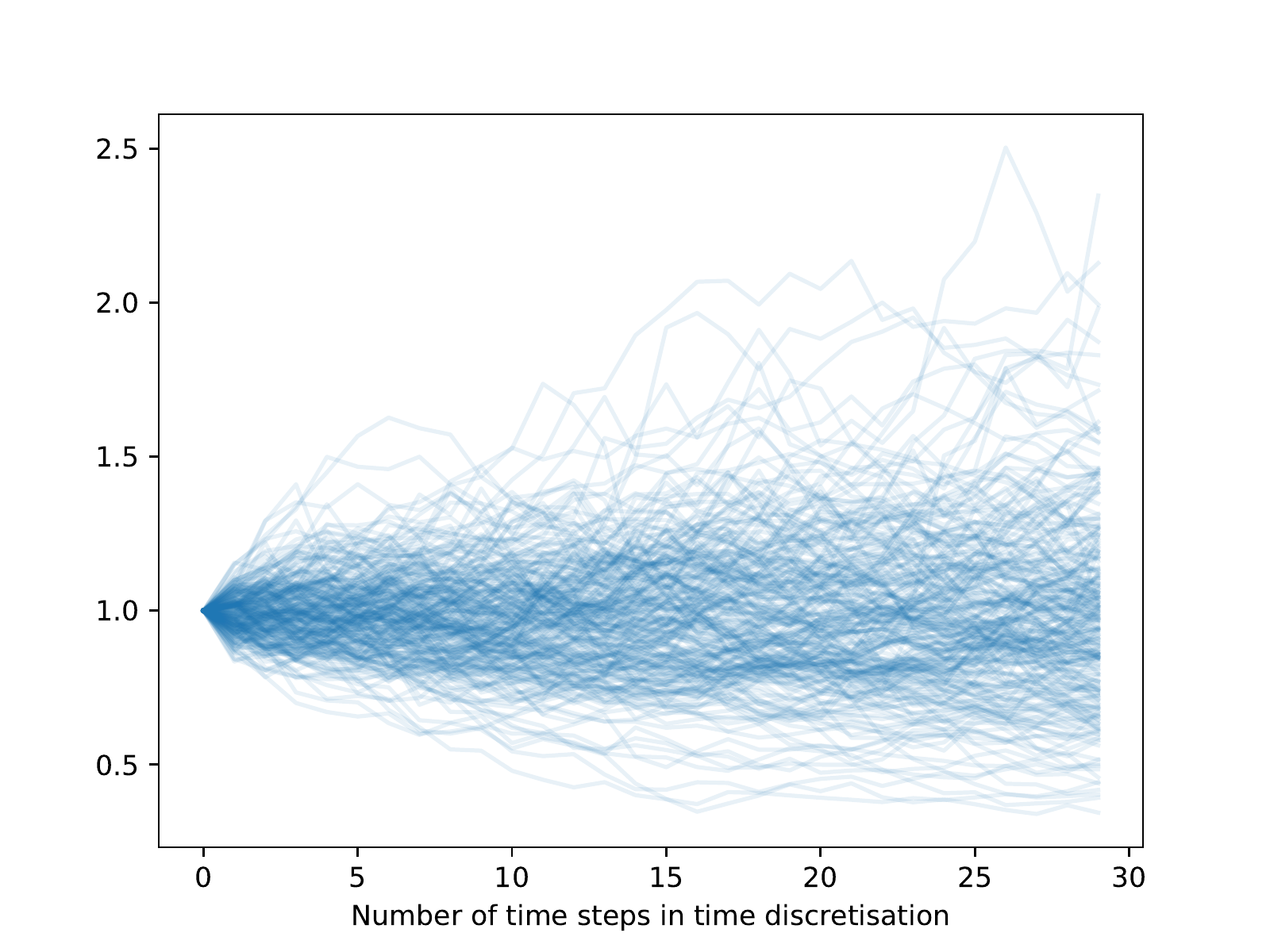}
  \caption{Real paths of new frequency}
  \label{fig:1}
\end{subfigure}
\begin{subfigure}{0.3\textwidth}
  \includegraphics[width=\linewidth]{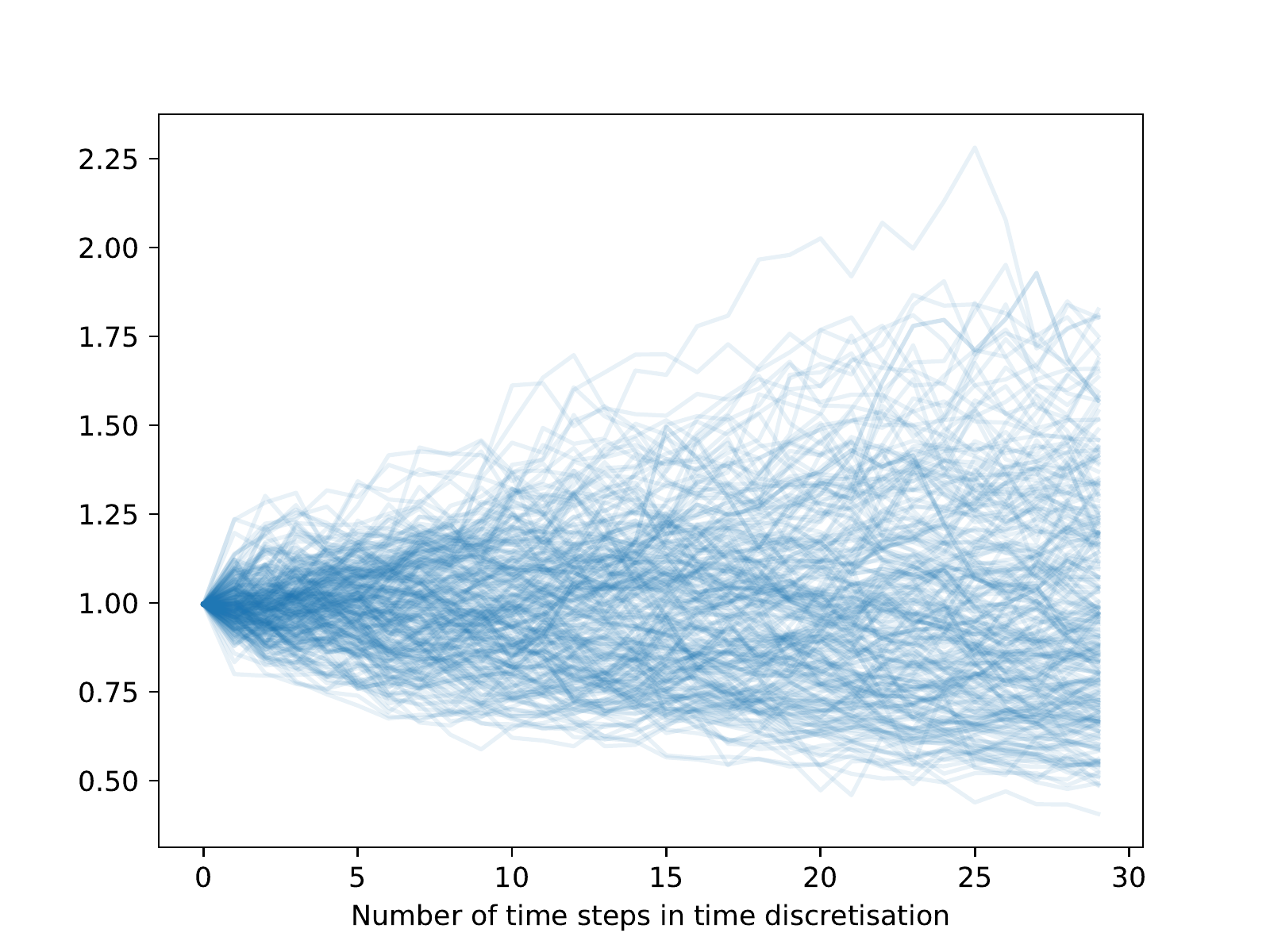}
  \caption{SigWGAN$+$LogsigRNN}
  \label{fig:2}
\end{subfigure}
\begin{subfigure}{0.3\textwidth}
  \includegraphics[width=\linewidth]{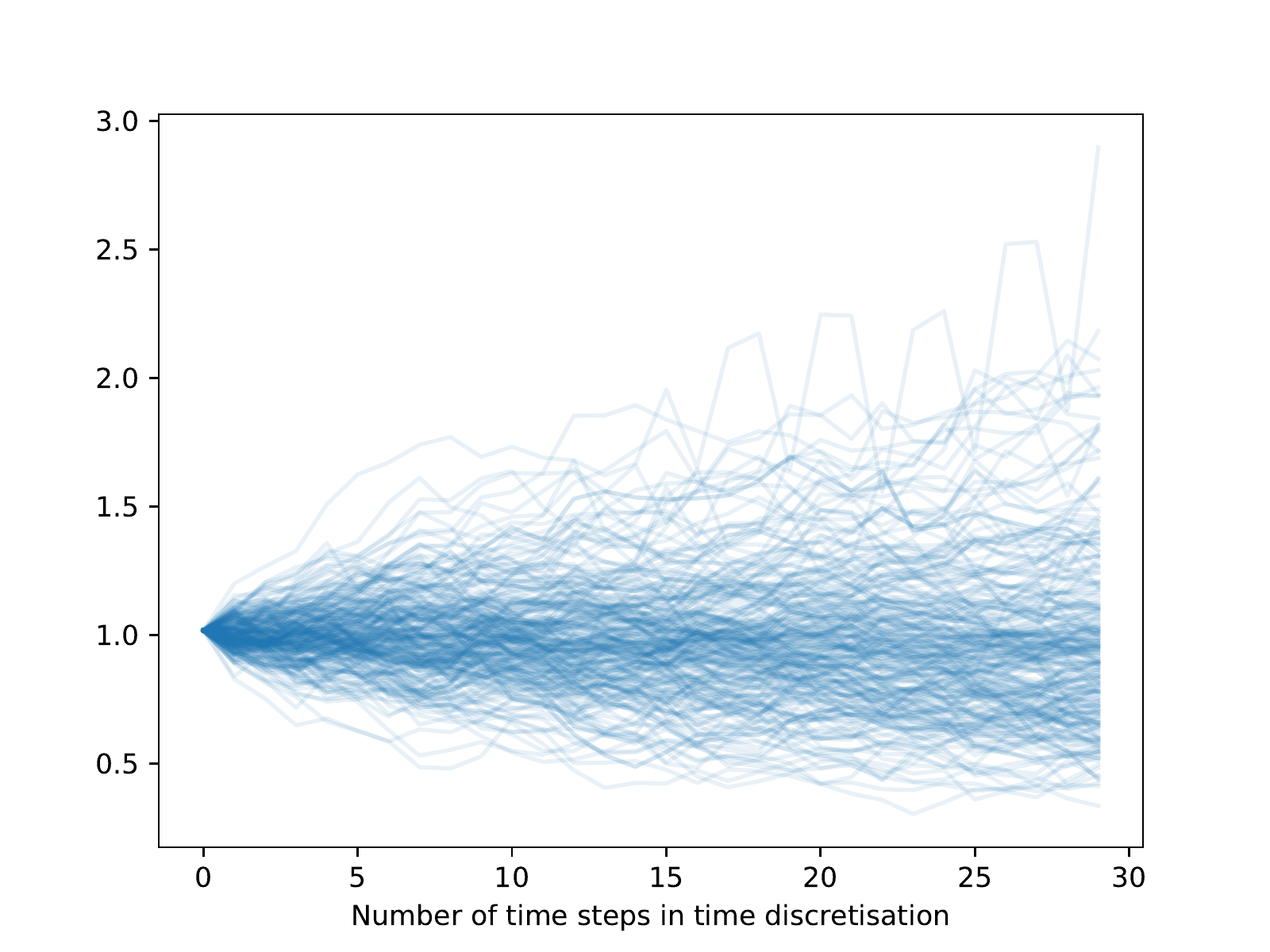}
  \caption{WGAN$+$LogSigRNN}
  \label{fig:3}
\end{subfigure}
\caption{From left to right, generated paths with new frequency (i.e. 30 time steps), generated paths using the trained LogSigRNN with Sig-W1 GAN on the old frequency (20 time steps), generated paths using the trained LogSigRNN with Wasserstein-GAN on the old frequency.}
\label{fig:robustness}
\end{figure}

We test the robustness of the trained generator by generating data streams with a different frequency than the frequency used for training and we test them on synthetic data generated from~\eqref{eq:bergomi} with the same new frequency. Namely, we generate data streams of 30 points on $[0,1]$, whilst training was performed on data streams of 20 points on $[0,1]$. Table~\ref{Table_RV_Robustness_Results} shows how in this new setting Sig-W1 GANs outperform GANs by two orders of magnitude. It is also verified qualitatively and visually by comparing the sample trajectories of the real paths and the synthetic path generated by the SigWGAN and WGAN using the same LogsigRNN generator in Figure~\ref{fig:robustness}.
\begin{table}[!ht]
\centering
\begin{tabular}{|l|c|c|c|c|}
\hline &\multicolumn{1}{c|}{W1} & \multicolumn{1}{c|}{Sig-W1 } \\ \hline
\centering{Data $(X_t)_{t}$} & $(S_t, v_{t})$   & $(S_t, v_{t})$  \\ \hline
\multicolumn{3}{|c|}{Correlation Metric(1e-4)}                     \\ \hline
LSTM & 358.84 & 8.32\\
\hline
LogsigRNN & 411.39 & 5.02\\
\hline
\end{tabular}
\caption{The test metrics of the trained models on two dimensional price and volatility data $(S_{t}, v_{t})_{t \in [0, T]}$. Models are trained on data streams sampled every $\frac{T}{20}$ units of time, and evaluated on data streams sampled every $\frac{T}{30}$ units of time. }\label{Table_RV_Robustness_Results}
\end{table}

\subsection{S$\&$P 500 and DJI Market Data}
We validate our proposed SigWGAN model on the empirical financial data. Here we choose daily spot-prices of the S\&P 500 and DJI from January 1st 2005 to June 1st 2020. This dataset is from the Oxford MAN Realized Library. We compute the log return series from the close price of both S\&P 500 and DJI, then apply rolling window to get 3856 samples of log return paths of $10$ time steps. We use $80\%$ samples as the training set and the rest $20\%$ as the test set.

To train the SigWGAN, we use the basepoint path augmentation, cumulative sum and the lead-lag transformation. According to Table \ref{Table_Stock_data_results}, for a fixed generator (LSTM/LogsigRNN), the SigWGAN improves the WGAN in terms of both the correlation metric and marginal distribution metrics. Besides, SigWGAN is more efficient than WGAN, which is illustrated by the half reduced training time when using the Logsig-RNN generator. Although the SigWGAN with the Logsig-RNN generator takes a bit longer to train than the other models, it can significantly reduce the correlation metric from $0.5982$ to $0.5031$ and the marginal distribution metric from $0.1752$ to $0.1304$ compared with the second best model. 

\begin{table}[!ht]
\centering
\begin{tabular}{|l|c|c|}
\hline Type&\multicolumn{1}{c|}{W1} & \multicolumn{1}{c|}{Sig-W1 } \\ \hline
\multicolumn{3}{|c|}{SigW1 Metric(1e-4)}                     \\ \hline
LSTM&5.0632&\textbf{2.7510}\\
\hline
LogsigRNN&5.6269&2.8161\\
\hline
\multicolumn{3}{|c|}{Correlation Metric(1e-1)}                     \\ \hline
LSTM&0.6422&0.5161\\
\hline
LogsigRNN&0.5982&\textbf{0.5031}\\
\hline
\multicolumn{3}{|c|}{Marginal Distribution Metric(1e-2)}                     \\ \hline
LSTM&0.2001&0.1513\\
\hline
LogsigRNN&0.1752&\textbf{0.1304}\\
\hline
\multicolumn{3}{|c|}{Training time(seconds)}                 \\ \hline
LSTM&0.138882&\textbf{0.103918}\\
\hline
LogsigRNN&0.472448&0.210008\\
\hline
\end{tabular}
\caption{The test metrics of the trained models on the log price of S$\&$P 500 index and  S$\&$P 500 and DJI index.}\label{Table_Stock_data_results}
\end{table}

\subsubsection{Correlation metric comparison}
\begin{figure}
     \centering
     \begin{subfigure}[b]{0.4\textwidth}
         \centering
         \includegraphics[width=\textwidth]{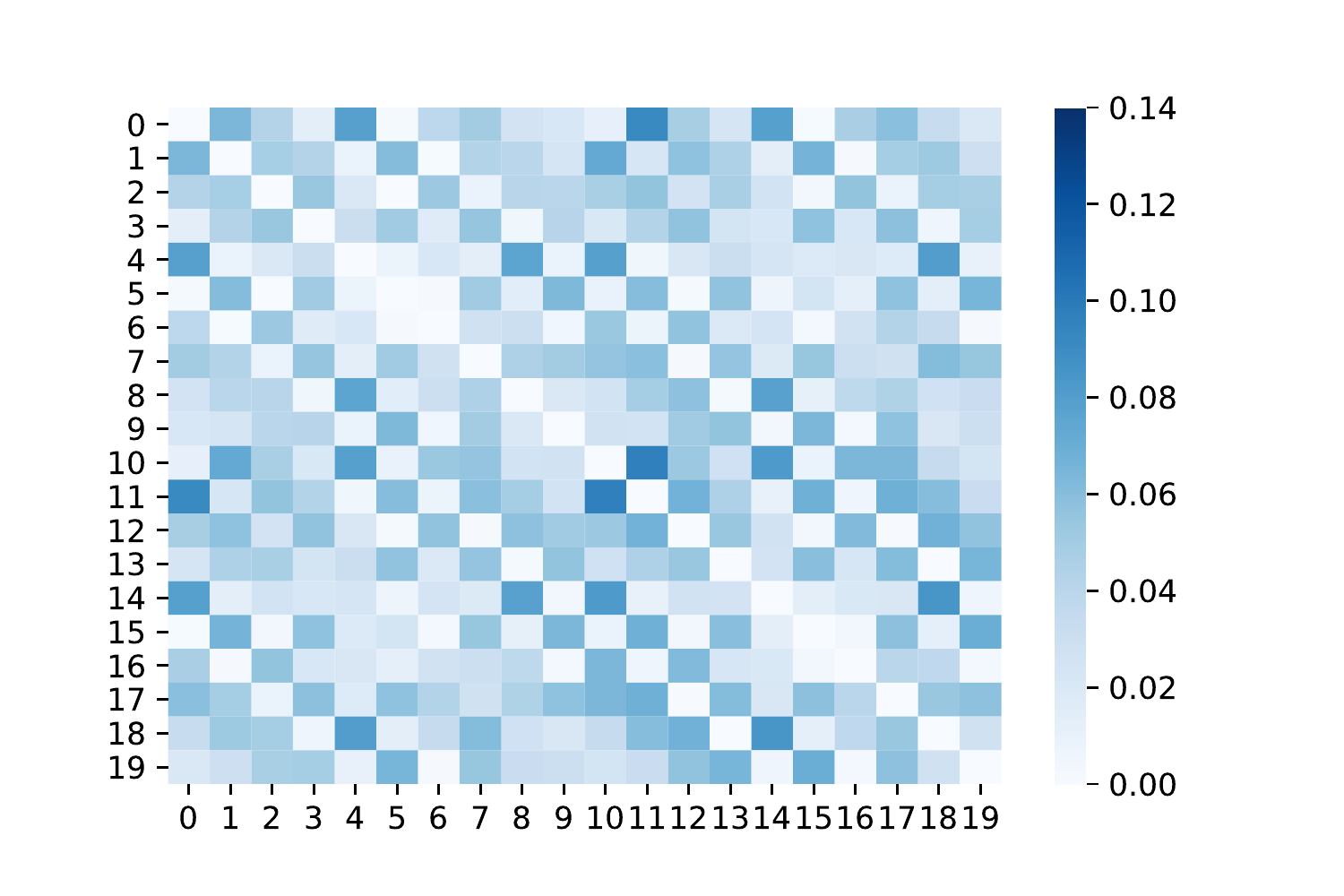}
         \caption{SigWGAN $+$ LogsigRNN}
         \label{fig:stock_corr_SigWGAN}
     \end{subfigure}
     \hfill
     \begin{subfigure}[b]{0.4\textwidth}
         \centering
         \includegraphics[width=\textwidth]{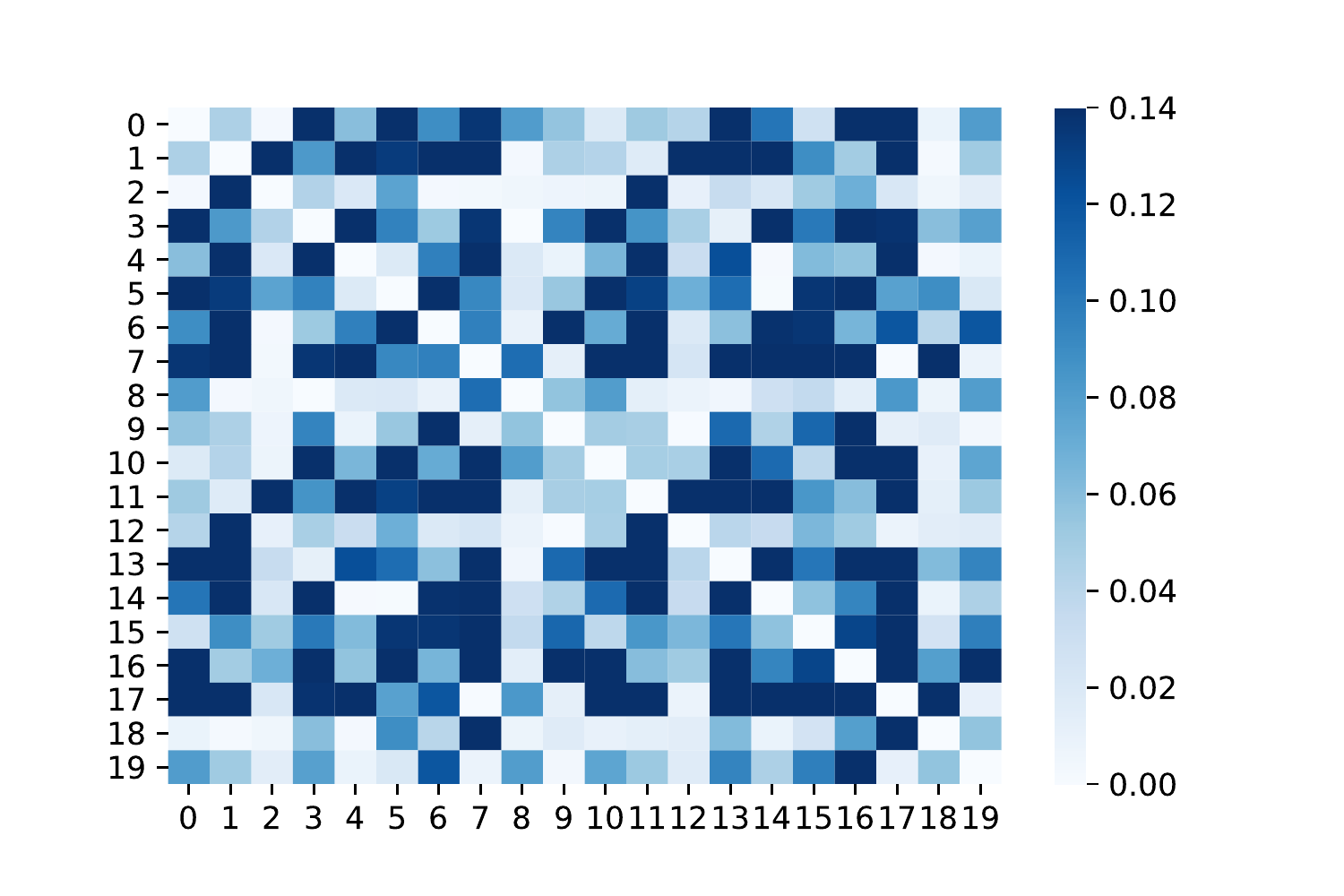}
         \caption{WGAN $+$ LogsigRNN  }
         \label{fig:stock_corr_WGAN}
     \end{subfigure}
        \caption{The correlation error of synthetic data generated by LogsigRNN using SigWGAN and WGAN resp. for stock data.}
        \label{fig:stock_corr}
\end{figure}

The error plot of the correlation matrix in Figure \ref{fig:stock_corr} clearly shows that the combination of the Logsig-RNN and SigW1 metric are able to capture the temporal and spatial dependency of the stock data best. However, the error plot of correlation matrix of Logsig-RNN and W1 metric shows more dark areas. Hence the combination of Logsig-RNN and W1 metric struggles to the temporal correlation of the volatility data.

\clearpage

\bibliographystyle{unsrt}
\bibliography{myref}  

\begin{thebibliography}{10}

\bibitem{assefagenerating}
Samuel Assefa, Danial Dervovic, Mahmoud Mahfouz, Tucker Balch, Prashant Reddy,
  and Manuela Veloso.
\newblock Generating synthetic data in finance: opportunities, challenges and
  pitfalls.

\bibitem{bellovin2019privacy}
Steven~M Bellovin, Preetam~K Dutta, and Nathan Reitinger.
\newblock Privacy and synthetic datasets.
\newblock {\em Stan. Tech. L. Rev.}, 22:1, 2019.

\bibitem{cuchiero2020generative}
Christa Cuchiero, Wahid Khosrawi, and Josef Teichmann.
\newblock A generative adversarial network approach to calibration of local
  stochastic volatility models.
\newblock {\em Risks}, 8(4):101, 2020.

\bibitem{Buehler2020data}
Hans Buehler, Blanka Horvath, Terry Lyons, Imanol~Perez Arribas, and Ben Wood.
\newblock A data-driven market simulator for small data environments, 2020.

\bibitem{Koshiyama}
Adriano Koshiyama, Nick Firoozye, and Philip Treleaven.
\newblock Generative adversarial networks for financial trading strategies
  fine-tuning and combination.
\newblock {\em Quantitative Finance}, 0(0):1--17, 2020.

\bibitem{gierjatowicz2020robust}
Patryk Gierjatowicz, Marc Sabate-Vidales, David Siska, Lukasz Szpruch, and Zan
  Zuric.
\newblock Robust pricing and hedging via neural sdes.
\newblock {\em Available at SSRN 3646241}, 2020.

\bibitem{Buehler2019}
Magnus Wiese, Lianjun Bai, Ben Wood, and Hans Buehler.
\newblock Deep hedging: learning to simulate equity option markets.
\newblock {\em Available at SSRN 3470756}, 2019.

\bibitem{QuantGAN}
Magnus Wiese, Robert Knobloch, Ralf Korn, and Peter Kretschmer.
\newblock Quant gans: deep generation of financial time series.
\newblock {\em Quantitative Finance}, 20(9):1419--1440, 4 2020.

\bibitem{Goodfellow}
Ian Goodfellow, Jean Pouget-Abadie, Mehdi Mirza, Bing Xu, David Warde-Farley,
  Sherjil Ozair, Aaron Courville, and Yoshua Bengio.
\newblock Generative adversarial nets.
\newblock In Z.~Ghahramani, M.~Welling, C.~Cortes, N.~Lawrence, and K.~Q.
  Weinberger, editors, {\em Advances in Neural Information Processing Systems},
  volume~27. Curran Associates, Inc., 2014.

\bibitem{arribas2020sig}
Imanol~Perez Arribas, Cristopher Salvi, and Lukasz Szpruch.
\newblock Sig-sdes model for quantitative finance.
\newblock {\em ICAIF 2020}, 2020.

\bibitem{kidger2020neural}
Patrick Kidger, James Morrill, James Foster, and Terry Lyons.
\newblock Neural controlled differential equations for irregular time series.
\newblock {\em Conference on Neural Information Processing Systems}, 2020.

\bibitem{ni2020conditional}
Hao Ni, Lukasz Szpruch, Magnus Wiese, Shujian Liao, and Baoren Xiao.
\newblock Conditional sig-wasserstein gans for time series generation.
\newblock {\em arXiv preprint arXiv:2006.05421}, 2020.

\bibitem{liao2021logsig}
Shujian Liao, Terry Lyons, Weixin Yang, Kevin Schlegel, and Hao Ni.
\newblock Logsig-rnn: a novel network for robust and efficient skeleton-based
  action recognition.
\newblock {\em Accepted by British Machine Vision Conference}, 2021.

\bibitem{morrill2020generalised}
James Morrill, Adeline Fermanian, Patrick Kidger, and Terry Lyons.
\newblock A generalised signature method for multivariate time series feature
  extraction.
\newblock {\em arXiv preprint arXiv:2006.00873}, 2020.

\bibitem{hambly2010uniqueness}
Ben Hambly and Terry Lyons.
\newblock Uniqueness for the signature of a path of bounded variation and the
  reduced path group.
\newblock {\em Annals of Mathematics}, pages 109--167, 2010.

\bibitem{boedihardjo2015uniqueness}
Horatio Boedihardjo and Xi~Geng.
\newblock The uniqueness of signature problem in the non-markov setting.
\newblock {\em Stochastic Processes and their Applications},
  125(12):4674--4701, 2015.

\bibitem{chevyrev2016characteristic}
Ilya Chevyrev and Terry Lyons.
\newblock Characteristic functions of measures on geometric rough paths.
\newblock {\em Annals of Probability}, 44(6):4049--4082, 2016.

\bibitem{mazumdar2019finding}
Eric~V Mazumdar, Michael~I Jordan, and S~Shankar Sastry.
\newblock On finding local nash equilibria (and only local nash equilibria) in
  zero-sum games.
\newblock {\em arXiv preprint arXiv:1901.00838}, 2019.

\bibitem{lin2020gradient}
Tianyi Lin, Chi Jin, and Michael Jordan.
\newblock On gradient descent ascent for nonconvex-concave minimax problems.
\newblock In {\em International Conference on Machine Learning}, pages
  6083--6093. PMLR, 2020.

\bibitem{daskalakis2018limit}
Constantinos Daskalakis and Ioannis Panageas.
\newblock The limit points of (optimistic) gradient descent in min-max
  optimization.
\newblock {\em arXiv preprint arXiv:1807.03907}, 2018.

\bibitem{daskalakis2017training}
Constantinos Daskalakis, Andrew Ilyas, Vasilis Syrgkanis, and Haoyang Zeng.
\newblock Training gans with optimism.
\newblock {\em arXiv preprint arXiv:1711.00141}, 2017.

\bibitem{mertikopoulos2018cycles}
Panayotis Mertikopoulos, Christos Papadimitriou, and Georgios Piliouras.
\newblock Cycles in adversarial regularized learning.
\newblock In {\em Proceedings of the Twenty-Ninth Annual ACM-SIAM Symposium on
  Discrete Algorithms}, pages 2703--2717. SIAM, 2018.

\bibitem{farnia2020gans}
Farzan Farnia and Asuman Ozdaglar.
\newblock Gans may have no nash equilibria.
\newblock {\em arXiv preprint arXiv:2002.09124}, 2020.

\bibitem{lucic2017gans}
Mario Lucic, Karol Kurach, Marcin Michalski, Sylvain Gelly, and Olivier
  Bousquet.
\newblock Are gans created equal? a large-scale study.
\newblock {\em arXiv preprint arXiv:1711.10337}, 2017.

\bibitem{chevyrev2018signature}
Ilya Chevyrev and Harald Oberhauser.
\newblock Signature moments to characterize laws of stochastic processes.
\newblock {\em arXiv preprint arXiv:1810.10971}, 2018.

\bibitem{liao2019learning}
Shujian Liao, Terry Lyons, Weixin Yang, and Hao Ni.
\newblock Learning stochastic differential equations using rnn with log
  signature features.
\newblock {\em arXiv preprint arXiv:1908.08286}, 2019.

\end{thebibliography}






\end{document}